\documentclass[a4paper,fleqn]{cas-sc}

\usepackage[authoryear]{natbib}
\usepackage{graphicx}   
\usepackage{epstopdf} 

\def\tsc#1{\csdef{#1}{\textsc{\lowercase{#1}}\xspace}}
\tsc{WGM}
\tsc{QE}
\tsc{EP}
\tsc{PMS}
\tsc{BEC}
\tsc{DE}

\begin{document}
\let\WriteBookmarks\relax
\def\floatpagepagefraction{1}
\def\textpagefraction{.001}
\shorttitle{BERT and Hierarchical LSTMs for Visual Storytelling}
\shortauthors{Jing Su et~al.}

\title [mode = title]{BERT-hLSTMs: BERT and Hierarchical LSTMs for Visual Storytelling}                      
\author[1,2]{Jing Su}

\address[1]{Guangdong University of Technology, Panyu District, Guangzhou 510006, China}
\address[2]{Guangdong Ocean University, Mazhang District, Zhanjiang 524088, China}

\author[1]{Qingyun Dai}

\author[3]{Frank Guerin}
\address[3]{University of Surrey, Guildford, Surrey GU2 7XH, United Kingdom}

\author[4]{Mian Zhou}
\cormark[1]
\ead{zhoumian@tjut.edu.cn}

\address[4]{Tianjin University of Technology, Xiqing District, Tianjin 300384, China}

\cortext[cor1]{Corresponding author \, at : Tianjin University of Technology}

\begin{abstract}
Visual storytelling is a creative and challenging task, aiming to automatically generate a story-like description for a sequence of images. The descriptions generated by previous visual storytelling approaches lack coherence because they use word-level sequence generation methods and do not adequately consider sentence-level dependencies.
To tackle this problem, we propose a novel hierarchical visual storytelling framework which separately models sentence-level and word-level semantics.
We use the transformer-based BERT to obtain embeddings for sentences and words.
We then employ a hierarchical LSTM network: the bottom LSTM receives as input the sentence vector representation from BERT, to learn the dependencies between the sentences corresponding to images, and the top LSTM is responsible for generating the corresponding word vector representations, taking input from the bottom LSTM. 
Experimental results demonstrate that our model outperforms most closely related baselines under  automatic evaluation metrics BLEU and CIDEr, and also show the effectiveness of our method with human evaluation. 

\end{abstract}

\maketitle
\section{Introduction}
Storytelling is one of the oldest known human activities for sharing narratives \citep{Miller2005}. Traditionally, storytelling is a way to educate, inculcate morals, preserve culture and instill advice. Recent  AI research has begun to tackle a storytelling task  known as visual storytelling: a machine is being used to automatically generate a sequence of coherent sentences (i.e., a story) for an ordered image sequence \citep{Cho2014Learning,Huang2016,Yulicheng2017}.

Inspired by the successful use of deep learning in machine translation \citep{Bahdanau2014Neural, transformer2017} and image captioning \citep{Karpathy2015Deep, Xu2015Show}, visual storytelling has attracted more attention in the field of vision and language. In contrast to image captioning where the descriptions of an individual image are generated automatically, visual storytelling is a more complicated and challenging task due to not only recognizing various objects and relationships within images but also learning the dependencies between images. Therefore, it is still an open research question: how to generate accurate and coherent story-like descriptions for sequential images?

Since it is difficult to construct a dataset for visual storytelling from scratch, the data were mainly extracted from the web in the initial visual storytelling research \citep{Kim2014, Sigurdsson2016}. Furthermore, photo streams and blog posts crawled from the web were collected and used to achieve story-based semantic summarisation \citep{KimSigal2015}. Subsequently, a dedicated dataset for visual storytelling known as VIST \citep{Huang2016} was released. The dataset contains stories where in each story, a group of five images was annotated with five corresponding descriptions using Amazon Mechanical Turk. The dataset makes it convenient to directly model the relationship between visual concepts or between visual imagery and typical activities.

Existing visual storytelling approaches can be divided into two categories: vision-based approaches and text-based approaches. Vision-based approaches mainly reconstruct sequential images or frames according to a storyline or plot. For example, \cite{Kim2014} exploit the data from Web community photos to formulate the photo selection as a sparse time-varying directed graph. \cite{Sigurdsson2016} summarise the photo album with an ordered collection of images by a Skipping Recurrent Neural Network (S-RNN). Next, \cite{SortStory2016} present multiple approaches by position and order predictions to sort a jumbled set of aligned image-caption pairs. All these approaches use the relationship between the images to sort image in sequences.

For text-based approaches, language models~\citep{Huang2016,Yulicheng2017,li2019stable} are typically exploited to generate story-like descriptions for sequential images or frames. For instance, \citet{Huang2016} introduce the first sequential vision-to-language dataset(VIST) and employ Gated Recurrent Units (GRUs) \citep{Cho2014Learning} based on an encoder and decoder mechanism for the task of visual storytelling. \cite{Yulicheng2017} propose a hierarchically attentive Recurrent Neural Network to address the visual storytelling task by firstly selecting the most representative photos and secondly composing a natural language story for the album. 

The vision and text approaches mentioned above mostly focus on summarizing or sorting sequential images or frames, and are a more shallow task than what is normally meant by  ``storytelling'' in everyday speech; however efforts to approach true storytelling are growing more recently. Existing approaches to visual storytelling have some limitations due to only considering the word-level representation by the recurrent neural net (RNN); this leads to difficulty in learning the relations between sentences and generating more coherent descriptions. To tackle the problem, we introduce a hierarchical LSTM (Long Short Term Memory), where one LSTM deals with the words and their dependencies, and the other LSTM deals with sentences and their dependencies. We propose a novel end-to-end model with a straightforward network structure called BERT-hLSTMs to generate story-like descriptions for sequential images. 

We mainly have the following three contributions: (1) We have exploited a pre-trained Bidirectional Encoder Representations from Transformer (BERT) model \citep{BERT2018} to obtain sentence representations and word representations of the dataset which can efficiently enrich the meaning of sentences. (2) Hierarchical LSTMs (hLSTMs) have been used to learn the relations between sequential images and corresponding descriptions and generate more coherent descriptions for sequential images. (3) We have evaluated the performance of our proposed approach on the the public visual storytelling dataset (VIST) \citep{Huang2016}. Experimental results show that our proposed model outperforms most closely related baselines under the automatic metrics BLEU and CIDEr, and performs well in human evaluation; it can generate more consistent descriptions and efficiently learn the dependencies between sentences corresponding to images.

The rest of this paper is organized as follows. Firstly, we introduce the related works on visual storytelling in Section 2. Secondly, we review our framework and describe the details of our proposed method in Section 3. Then, we show the experimental results and evaluate our BERT-hLSTMs model on the VIST dataset in Section 4. Finally, Section 5 gives conclusions and future work.

\section{Related Work}
Visual Storytelling is at the intersection of language generation and computer vision, and goes beyond single image description, to sequential images. The technique builds on Image/Video Captioning, combined with Visual Summarisation/sort.

\subsection{Image and Video Captioning}  The successes in machine translation using RNN \citep{ Cho2014Learning, RNN2014,Bahdanau2014Neural} catalysed the explosion of research in image/video captioning. Image/video captioning based on Convolutional Neural Networks (CNN) and Recurrent Neural Networks (RNN) or variants \citep{Vinyals2014Show, Karpathy2015Deep, Donahue2014} has made great progress. \cite{Vinyals2014Show} present an encoder-decoder model which combines visual representation with a CNN \citep{Simonyan2014Very} and textual representation with a Recurrent Neural Network (RNN). \cite{Karpathy2015Deep} propose a CNN and Bidirectional RNN based model which can align segment regions in an image to the corresponding textual section. \cite{Mao2014a} propose a multimodal RNN model which additionally leverages a multimodal layer to combine a CNN model and language model together.

Subsequently, Long Short Term Memory (LSTM) with an attention mechanism \citep{Xu2015Show, Zhu2015Aligning, You2016Image,Lu16,su2018generating,li2019dual} has been widely used and proved to be effective for image captioning. For instance, \cite{Xu2015Show} explore two kinds of attention mechanism for generating image descriptions, whereas \cite{You2016Image} exploit a selective semantic attention mechanism for the same task.

The approaches based on CNN-RNN or CNN-LSTM with attention are also extensively applied to video captioning \citep{Venugopalan2014Translating, Venugopalan2015Sequence, Yao2015Describing}. \cite{Venugopalan2015Sequence} propose a novel LSTM model to generate captions for videos. The model associates a sequence of video frames with a sequence of words in order to generate descriptions of the event in the video clip on a standard set of YouTube videos and two movie description datasets (M-VAD and MPII-MD). \cite{Yao2015Describing} use a global structure with a temporal attention mechanism among video frames to address the video captioning task. This temporal attention mechanism based on a soft-alignment approach generates more relevant context for a predicted word due to dynamically focusing on key frames.

\subsection{Visual Summarisation/Sort} Visual summarisation mainly selects key frames or images in a sequence to compose a storyline. \cite{khosla2013} use unsupervised learning or intuitive criteria to pick salient frames. \cite{PhotoSequencing2014} proposed a method for photo-sequencing, to temporally order a set of still images from uncalibrated cameras. Their approach detects sets of corresponding static and dynamic feature points between images, with static features being used to determine the epipolar geometry between pairs of images, and dynamic feature voting for the temporal order of the images. \cite{Gong2014} and \cite{Zhang2016} proposed a new model to learn from human-created summaries. 

Storyline graphs are directed graphs which have been used to capture chronological or causal relations between image clusters \citep{Kim2014}. Somewhat related is \citet{Sigurdsson2016} which implicitly learns the same kind of graph within its Recurrent Neural Network.
\citet{Bosselut2016} cluster images and text from web albums of a single concept or event (e.g. wedding), to learn the hierarchical events that make up these scenarios (i.e. the storyline). More recently, \cite{Choi2017} proposed to customise the selection based on a given  user-specific text description, and hence select semantically relevant video segments. 

Some other works in vision \citep{Basha2014,Pickup2014} also temporally order images typically by finding correspondences between multiple images of the same scene using geometry-based approaches. Similarly, \cite{Choi2016} leverage dense optical flow features and a patch matching algorithm to define metrics on scene dynamics and coherency and employ plot analysis to compose a video for a given set of multiple video clips. 
\cite{Ramanathan2015} deal with sampled frames from a video and attempt to learn a temporal embedding of video frames in complex events. 
\cite{SortStory2016} combine text-based and image-based features and use  unary and pairwise predictions to sort a jumbled set of aligned image-caption pairs into a sequence which forms a coherent story.

\subsection{Visual Storytelling } Visual storytelling takes as input a sequence of images and attempts to describe a coherent story for them, in text. While the input, and the training sets could be formally defined, the goal is not to simply produce the same label (text) as a test set; instead it is normally described as producing a story which a human judge would consider to be a good story, although in practice automatic evaluation metrics are heavily used as a proxy for human judgement.
Some of the works in this subsection overlap with the previous subsection in that they select images in addition to telling the story. Work in this area can be clustered in three groups: (1) direct deep learning without intermediates; (2) pipelines which use some intermediate data to help the storytelling; (3) reinforcement learning with clever reward functions.

\subsubsection{ Direct deep learning without intermediates}\label{without}
\citet{ImageStream2015} design a multimodal architecture called coherent recurrent convolutional network (CRCN), which consists of convolutional neural networks, bidirectional recurrent neural networks, and an entity-based local coherence model. 
The input is images and story sentences and the output is a score for compatibility  between image stream and story.
They train their model on sequences of sentences accompanying image streams from online natural blog posts. Unlike later works, \citet{ImageStream2015} do not decode text output from a hidden state, but rather choose sentences from existing blog posts, based on their score computed by the trained network.

While earlier works obtained training data from e.g. blog posts, more recently \cite{Huang2016} introduced the first Visual Storytelling dataset (VIST)  for sequential vision-to-language storytelling. This has subsequently been widely adopted.
\cite{Huang2016} also provided an encoder-decoder RNN baseline for the task.
\cite{Yulicheng2017} use a hierarchically-attentive Recurrent Neural Net (GRU-RNNs) to encode the album photos, select representative photos, and compose stories for photo albums in the VIST dataset.  
\cite{Yulicheng2017} and \citet{wang2019} additionally tackle the problem of selecting photos from an ordered stream. \cite{Yulicheng2017} compute a photo's attention and pick photos with the highest probability of inclusion, while \citet{wang2019}  introduce a scene encoder which can determine when the current photo describes the start of a new scene. Apart from this, \citet{wang2019} is an example of an end-to-end trainable encoder-decoder approach, using GRUs.

\subsubsection{ Exploiting intermediate data or structures}\label{interm}

Automatically learning to map from image sequences  to output stories is very challenging with no guidance, hence some approaches try to introduce some intermediate representation or data to help.
A simple approach is taken by \citet{Nahian2019}, which encodes images and their associated text captions (from the VIST dataset) by separate encoders, and combines them, before decoding into the story sentences. Otherwise \citet{Nahian2019} is a fairly straightforward encoder-decoder architecture.
Other works try to extract some semantic information from the images without simply using the caption given in the dataset.
\citet{li2019} learn association rules between images and the topics in human generated captions. They can then use these to (speculatively) extract the topic or event out of given visual input (e.g. stadium, military). This extra input is then included in an encoder-decoder architecture using GRUs.
\citet{Zhang2020} similarly learn words called  ``anchor words'' which capture important semantics and help the narrative generation.
\citet{Huang2019} generate a semantic concept or topic word for each image in sequence, and use this to assist the narrative generation; they train using reinforcement learning.
\citet{li2019t} use text from similar stories as an additional input to help in the generation of the current stories. Similar stories are found based on image similarity.

In contrast to those approaches adding text input, \citet{wangaaai2020} extract a scene graph from each image, as an intermediate representation to help generation. They then process this with a Graph Convolution Network (GCN), followed by a  Temporal
Convolution Network (TCN) to capture temporal relationships across images. Visual feature maps are also fused with this representation.
Decoding is done by GRU with an attention mechanism, to produce story sentences.

\subsubsection{ Reinforcement Learning}\label{RL}

Reinforcement Learning has been applied to visual storytelling, as it has been successfully applied to captioning before. Typically the encoder-decoder models are the same as in the previous sub-section, but the way they are optimised is very different (by reinforcement learning). In general reinforcement learning allows for much more flexibility in how an objective can be defined, when compared to the standard maximum likelihood estimation of the previous subsections.
In visual storytelling it is challenging to find a good reward function, because, unlike image captioning, there is a much wider variety of potential stories that are all good. If one borrows metrics such as BLEU \citep{BLEU2002}, METEOR \citep{meteor}, or CIDEr \citep{CIDEr2014} the rewards can be too sparse, or may lead the system to game the metric
\citep{wang-etal-2018-metrics}. Therefore much of the work in this area is devoted to finding clever reward functions.

The reward is in general a sophisticated function of the story output and the visual features.
\citet{wang-etal-2018-metrics} design an adversarial training for the reward function where a Boltzmann distribution aims to maximise similarity with the training data and minimise similarity with generated `fake' stories.
The generator (policy) on the other hand tries to maximise similarity with the Boltzmann distribution. Their system is called  Adversarial
REward Learning (AREL).
\citet{Huang2019} use a reward function that balances the requirement to be relevant to the semantic concepts of the story, with the requirement to follow its language model. The semantic concepts come from the intermediate data mentioned in the previous subsection. 

From the above works we are in the category of works in Sec.~\ref{without} which have no intermediate data like topic words, or scene graph. We  have some further similarity with \citet{wang2019} because it uses attention on the photos.
However, none of the above works separately models the text semantics at sentence-level also and word-level; this is something we are introducing to improve the coherence of the output story. This idea of modelling both word and sentence level semantics has already been successful in text modelling tasks outside of visual storytelling \citep{li-etal-2015-hierarchical,Serban,zhang2018sentencestate}.

\section{Vision to Language Model}

We propose a novel visual storytelling framework called BERT-hLSTMs combining the superior BERT model and a hierarchical LSTMs model for automatically generating  story-like descriptions of sequential images. 
As illustrated in Figure~\ref{FIG:story_caption}, the proposed framework has three major components:

Firstly, like previous works, we use a Convolutional Neural Network (CNN) in the encoder of our proposed model, using the VGG16 network \citep{Simonyan2014Very}. However, the difference is that from this we take three different types of visual features as inputs to the next stage:

\begin{figure}[pos=tb]
	\centering
		\includegraphics[scale=0.6]{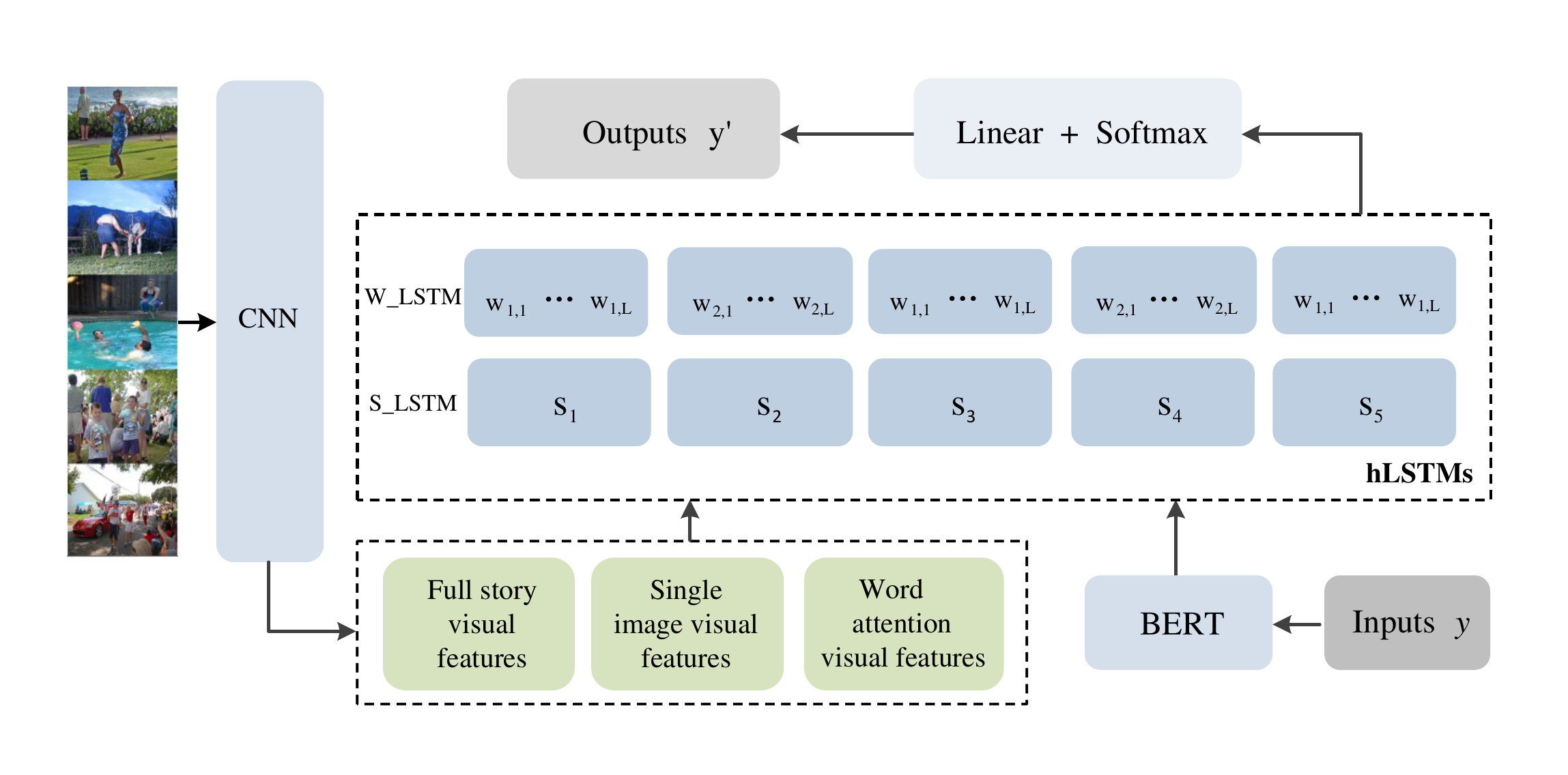}
	\caption{The architecture of our BERT-hLSTMs model. It consists of three components: CNN, BERT ans hLSTMs. The CNN component is used to extract the features of sequential images. The BERT component is used to embed the sentences and words in the corresponding descriptions of sequential images. The hLSTMs component is responsible for learning the relations between generated descriptions.}
	\label{FIG:story_caption}
\end{figure}

\begin{enumerate}
    \item Full-story visual features combine the features of all five images in a story; these are used to  initialise the hidden state of the LSTMs.
    \item Single-image visual features, for each image, are used individually by each sentence level LSTM (S\_LSTM) that needs to write the description for that image.
    \item Word-attention visual features used by the word level LSTM (W\_LSTM) are features of the image selectively attended to because of the current word.
\end{enumerate}

\begin{figure}[pos=t]
	\centering
		\includegraphics[scale=0.65]{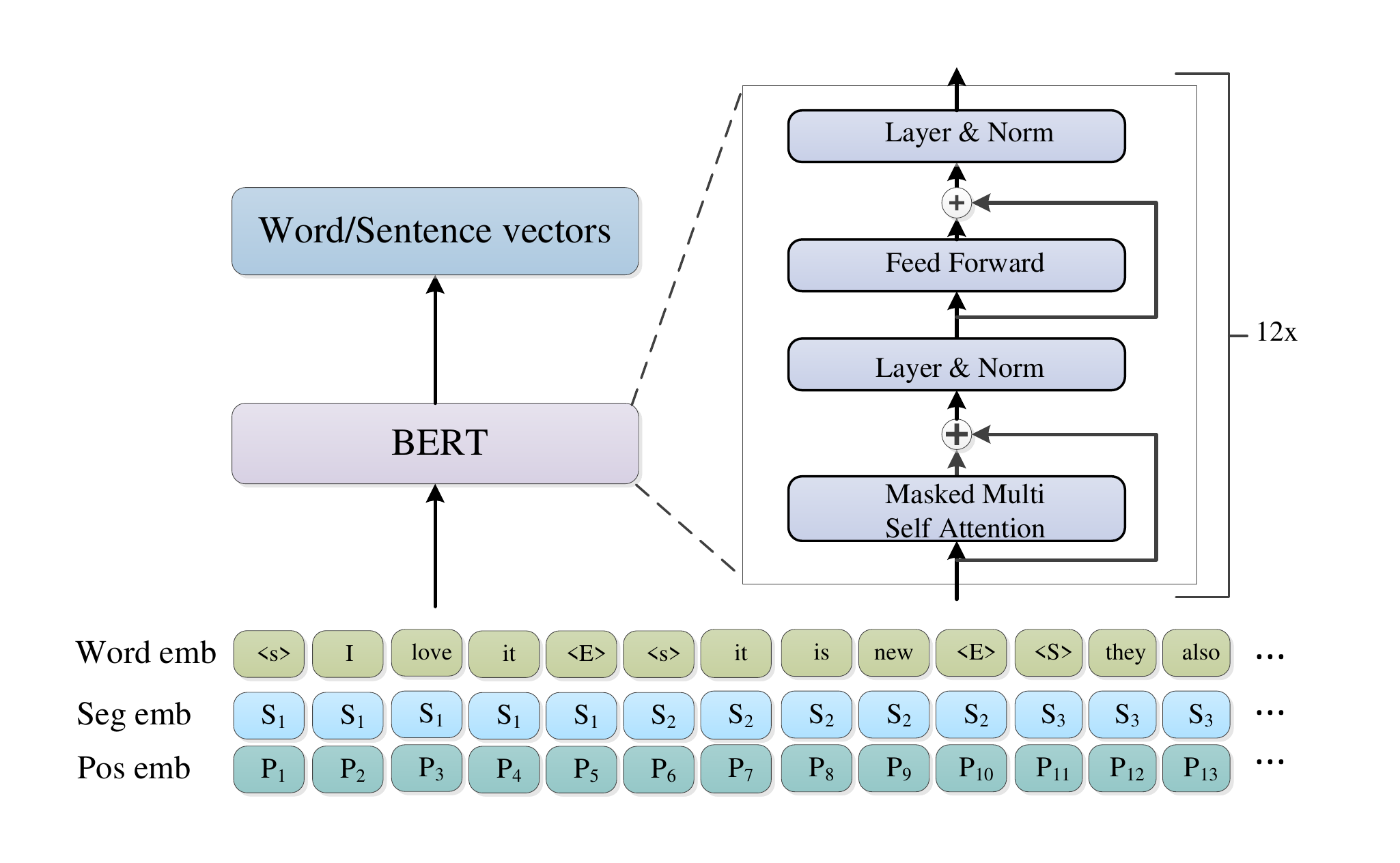}
	\caption{Illustration of BERT embedding. }
	\label{FIG:bert}
\end{figure}

Secondly, BERT is conceptually simple and empirically powerful when used for word embedding; it has obtained new state-of-the-art results on several language tasks. Therefore, we introduce a pre-trained BERT \citep{BERT2018} for embedding the words and sentences, which can efficiently enrich the meaning of words and sentences in vector spaces and then perform fine-tuning to the word vector and sentence vector, by feeding them into the model. The architecture of the BERT model is a multi-layer bidirectional Transformer encoder \citep{transformer2017}, where the Transformer incorporates a self-attention mechanism and is mainly composed of multi-head attention, feed-forward layers and layer normalization. 
 
As shown in Figure~\ref{FIG:bert}, the BERT model is responsible for translating the information of the words into numerical representations. At the same time, we exploit the BERT model to directly produce sentence vector representations. Consequently, our model is capable of focusing on not only the word-level contextual information but also the sentence-level contextual information.

Finally, we use hierarchical LSTMs as a language model with  long-term sequence information to generate  story-like descriptions. The hierarchical LSTMs is a two-layer LSTM model: S\_LSTM layer and W\_LSTM layer. The first LSTM layer, S\_LSTM, is used to generate next sentence vector representation, conditioned on the previous sentence vector representation, in each time step. This learns the relations between images as well as sentences. The second LSTM layer,  W\_LSTM, is used to predict each word based on the sentence from the bottom S\_LSTM layer. The W\_LSTM layer exploits an attention mechanism similar to \citet{Xu2015Show} and fuses the sentence-level information and word-level information by taking the output of the S\_LSTM layer as the initial input of each sentence in the W\_LSTM layer.

\subsection{Visual Features Extraction}
Given a sequence of images where the number of images is N, we use the final convolution layer of the VGG16 network to extract the spatial feature maps of the sequential images as follows:
\begin{enumerate}
    \item Full-story visual features: The complete image feature output from the VGG16 final convolution, before transformation, is $x$,  where $x \in R^{N \times M \times D}$. $M$ is the number of feature vectors for each image in a story, with each feature vector being $D$-dimensional.
    We add a convolution layer with a weight matrix  having the same dimensions as that of the hidden unit in the LSTM model, and perform a dimensional transformation to gain the final features of all $N$ images in the model. This is the  initial input to the S\_LSTM and W\_LSTM hidden layer, while we employ  different single-image visual features after. 
    \item Single-image visual features $x_i$: features for each image $i$, are used individually by each sentence level LSTM (S\_LSTM) that needs to write the description for that image.
    \item Word-attention visual features: These are computed from the single-image visual features using an attention mechanism (like \citet{Xu2015Show}), and are then used by the word level LSTM (W\_LSTM). They are features of the image selectively attended to based on the current word.
\end{enumerate}

\subsection{Text Embedding}
A story is represented by $y = \{y_1,y_2,…, y_N\}$, which is a sequence of   descriptions of sequential images. $N$ is the the number of sentences. Each $y_i$ represents a sentence,  $y_i \in R^L$,  and $L$ is the number of words in a sentence. In order to obtain a high-quality embedding of the sentences and words in a story, we use the BERT model for sentence embedding and word embedding unlike earlier works which instead used Word2Vec or one-hot vectors. We  construct a sentence dictionary of all the sentences, as well as a word dictionary which contains all the words in the dataset. Then the pre-trained BERT model is adopted to embed each sentence and each word in the dictionaries. With the word embedding model, we get the sentence vectors of a story $s \in R^{N \times D}$ and word vectors of a story $w \in R^{N \times L \times D }$ where $D$ is equal to the dimension of the LSTM hidden unit. Finally, we perform fine-tuning by feeding sentence vector representations and word vector representations corresponding to the story descriptions into the hierarchical LSTMs.

\subsection{Story Generator}
In this subsection we explain our proposed approach for story-level generation. A novel hierarchical LSTMs model is adopted. The S\_LSTM layer is for sentence-level semantic generation with the corresponding sentence vector representation from the BERT model as the input. The W\_LSTM layer is for the word-level semantic generation, with initial input coming from the output of the S\_LSTM layer (as illustrated in Figure~\ref{FIG:story_caption}).

Firstly, our model uses the popular Long Short-Term Memory (LSTM) network with attention mechanism \citep{Xu2015Show} where the  output at each time step is conditioned on the current semantic context information and the previously generated hidden state. The structure of the LSTM with attention is shown as follows:

\begin{eqnarray}
&& i^t = \sigma(W_{xi}x^t+W_{hi}h^{t-1}+W_{zi}z^t+b_i) \nonumber\\
&& f^t = \sigma(W_{xf}x^t+W_{hf}h^{t-1}+W_{zf}z^t+b_f) \nonumber\\
&& o^t = \sigma(W_{xo}x^t+W_{ho}h^{t-1}+W_{zo}z^t+b_o) \nonumber\\
&& q^t = \mathit{tanh}(W_{xq}x^t+W_{hq}h^{t-1}+W_{zq}z^t+b_q) \nonumber\\
&& c^t = f^t \odot c^{t-1} + i^t \odot q^t \nonumber\\
&& h^t = o^t\odot \mathit{tanh}(c^t) \label{eq3}
\end{eqnarray}

Where $i^t$ represents input gates, $f^t$ represents forget gates, $o^t$ represents output gates and $c^t$ represents memory state. In addition, $q^t$ represents the updating information in the memory state $c^t$. $\sigma$ denotes the sigmoid activation function, $\odot$ denotes the element-wise multiplication, and $\mathit{tanh}$ indicates the hyperbolic tangent function. $W_{(\bullet)}$ and $b_{(\bullet)}$ are the parameters to be learned during training. Also $h^t$ is the hidden state at time step $t$ which will be used as an input to the LSTM unit at the next time step.

For simplicity, we define each LSTM unit mentioned above as:
 \begin{eqnarray} \label{eq:lstm}
&& h^t, c^t = \text{LSTM}([x^t,z^t], h^{t-1}, c^{t-1}) 
 \end{eqnarray}
 
LSTM$(\centerdot)$ represents the computing function of the LSTM unit. That is, the hidden state $h^t$ and the memory state $c^t$ at the time step $t$  depend on the current semantic context information, which is the concatenation of the current textual vector $x^t$ and the current dynamic vision vector $z^t$, the previous hidden state $h^{t-1}$ and previous memory state $c^{t-1}$. Then, for the S\_LSTM layer, our approach can be defined as below:

\begin{eqnarray}
&&h^o_s =W^o_s \phi(x) \nonumber\\
&&x^t_s = \psi(x) \nonumber\\
&&h^t_s, c^t_s = {\text{LSTM}_{sent}}([s^t,x^t_s], h^{t-1}_s, c^{t-1}_s) 
\end{eqnarray}

The initial value of the hidden state $h^o_s$ is obtained by employing a fully connected network to transform the full-story visual features $x$. $W^o_s$ are the parameters that need to be learned. $x^t_s$ denotes the visual feature vector at time step $t$ and is equal to the feature vector representation of the $t$-${th}$ image. 
$\phi$ denotes the mean function. $\psi$ denotes the operations of $x$ to obtain the features of the $i$-${th}$ image.
Here we define the input of the model: the semantic context information is composed of the current sentence vector representation $s^t$ and current visual feature representation $x^t_s$ at time step $t$. Therefore, the output of the model $h^t_s$ and the memory state $c^t_s$ at the time step $t$ are computed by feeding the current semantic context information, previous hidden state $h^{t-1}_s$ and previous memory state $ c^{t-1}_s $ into the function $\text{LSTM}_{sent}$.

For the W\_LSTM layer, we integrate the sentence-level context information and the word-level context information to predict the word of each sentence in a story in order. The sequence generation model can be formulated as below:

\begin{eqnarray}
&& h^o_t = W^o_t \, \phi(x^t_s)  \nonumber\\
&& \alpha^k_{t,j} = \text{softmax}(\Phi(h^{k-1}_t, x^t_s))  \nonumber\\
&& z^k_t = \sum_{j=1}^L (\alpha^k_{t,j}x_{t,j})   \nonumber\\
&& h^k_t, c^k_t = {\text{LSTM}_{word}}([h_s^t,z_t^k], h_t^{k-1}, c_t^{k-1}) \,\,(k=1) \nonumber\\
&& h^k_t, c^k_t = {\text{LSTM}_{word}}([w_t^{k-1},z_t^k], h_t^{k-1}, c_t^{k-1}) \,\,(k>1)   \label{eq4} 
\end{eqnarray}

Here, $x_t$ denotes the single-image visual features of the $t$-${th}$ image in a sequence. $k$ is the time step of the word sequence conditioned on the generated sentence vector from the S\_LSTM layer and ranges up to  the size of the sentence. 
 $\Phi$ denotes the linear projection function.
The weight $\alpha^k_{t,j}$ can be viewed as the attentive probability of the $j$-${th}$ location in the $t$-${th}$ image while predicting the next word of the $t$-${th}$ sentence at the time step $k$, and $z_t^k \in R^D$ is the word-attention visual features which enable the model to focus on the relevant portion of the visual features at the current moment.

In the W\_LSTM layer, $v^k_t$ $\in$ $R^D$ denotes the textual context representation of the $t$-${th}$ sentence at the time step k and has the same dimension as the hidden state of the LSTM unit. When the time step $k$ is equal to 1, $v^k_t$ is defined as the output $h^t_s$ of the S\_LSTM layer while $v^k_t$ is the word vector $w^{k-1}_t$ at the time step $k > 1$. Consequently, we can obtain the hidden state $h^k_t$ which also is the output of W\_LSTM layer, and the current memory state $c^k_t$ by the function $\text{LSTM}_{word}$ based on the inputs mentioned above (as seen in Eq.~\ref{eq4}).

The goal of our framework is to generate a coherent description for a given sequence of images. This can be obtained by maximizing the probability of the generated story-like descriptions $y$ given the visual features $x$ of sequential images and the model parameters $\theta$. Assuming that a generation model of visual storytelling produces each sentence in order, with each sentence containing $L$ words, the probability of generating story-like descriptions is given by the sum of the probabilities over the sentences and the probability of each sentence is the joint probability of a sequence of words. This is as shown in Eq.~\ref{eq:probability}
 
\begin{eqnarray} \label{eq:probability}
P(y_i|x,\theta) && = \prod_{k = 1}^{L} P(w_{i,k}|x,w_{i,1},w_{i,2},...,w_{i,k-1}) \nonumber\\
P(y|x,\theta) && = \sum_{i=1}^N P(y_i|x,\theta) 
\end{eqnarray}

where $w_{i,k}$ represents the $k$-${th}$ word of the $i$-${th}$ sentence and $L$ is the size of the sentence.

We employ  cross entropy loss to train the model. Therefore, The loss can be obtained by minimizing the sum of the log probability of each sentence in a sequence. The loss function is defined as Eq.~\ref{eq:loss}.

\begin{equation} \label{eq:loss}
L = -\sum_{i=1}^N \sum_{k = 1}^{L}logP(w_{i,k}|x,w_{i,1},w_{i,2},...,w_{i,k-1})
\end{equation} 

We choose the Adam optimizer to optimize the above objective function. At test time we adopt a beam search strategy to generate the final story-like descriptions. The hyper-parameter settings  will be introduced in the next section.

\section{Experiments}

We experiment with our BERT-hLSTMs model on the task of visual storytelling. In this section we describe the dataset used for this task, the baseline methods we compare with, and implementation details of our approach. Finally, we evaluate the performance of our proposed framework by comparing our results with the baseline methods across the standard evaluation metrics.

\noindent {\bf Dataset.} We evaluate our model by conducting experiments on the VIST dataset\footnote{\url{http://visionandlanguage.net/VIST/}} published by Microsoft \citep{Huang2016}, which contains 10,117 Flickr albums and 210,819 unique photos. VIST has two different types of the descriptions for the same set of images: descriptions (captions) of images-in-isolation (DII) and stories for images-in-sequence (SIS). DII mainly focuses on typical event patterns and has no dependencies between images in each sequence, while SIS has dependencies between sequential images and more abstract expressions which connect the literal descriptions to more abstract visual concepts. Here, we exploit the SIS descriptions for the experiment where each album contains 5 stories and each story has 5 corresponding  descriptions  for 5 images in a sequence. Each description is typically one sentence, but some can be more in rare cases. The stories in the same album may describe the same sequence of images or may describe  different sequences of images within the same album.

The sentences in the stories have a typical size of about 15 words. So we extract a subset of the dataset for our experiments where the size of each sentence is no longer than 15. The new dataset with the vocabulary size of 18,000 consists of 22,367 stories for training, 2,300 stories for validation and 2,300 for testing. 

\subsection{ Implementation} We implement our model based on the Tensorflow framework with Python. We use the VGG16 model as the encoder to extract the visual features. We extract the output of the final convolutional layer as the image intermediate feature. The dimension of this has the shape of $5 \times 14 \times 14 \times 512$, where 5 represents the number of images in a story, $14 \times 14$ is the height and the width of the intermediate feature map, and 512 is the channel number of the intermediate feature map. In our model, the first two visual feature representations are  the full-story visual features  and  the single-image visual features. We firstly add a convolutional network with  weights $w \in R^{512\times768}$ to turn the intermediate feature into a vector with  shape  5 $\times$ 14 $\times$ 14 $\times$ 768. For the single-image visual features, we transform it into a 5 $\times$ 196 $\times$ 768-dimensional matrix which can be viewed as an image spatial matrix with 14 $\times$ 14 grids and every part being represented by a 768-dimensional vector. For the full-story visual features, it is transformed into a 5 $\times$ 768-dimensional matrix by computing the sum of the 14 $\times$ 14 grids. Finally, we get a 5 $\times$ 768-dimensional full-story visual features vector and a 5 $\times$ 196 $\times$ 768-dimensional single-image visual features matrix. 

\begin{table}[tb]
\caption{The performance comparison between our approaches with the baselines on the VIST dataset.}\label{tbcomparison}
\begin{tabular*}{\tblwidth}{@{} LLL@{} }
\toprule
 \bf Method  & \bf BLEU & \bf CIDEr\\
\midrule
enc-dec (variant of \citet{Vinyals2014Show} extended for  image sequences)  &  19.58  & 4.65 \\
enc-attn-dec (variant of \citet{Xu2015Show} extended for  image sequences)  &  19.73 &  4.96 \\
h-attn \citep{Yulicheng2017}  &  20.53 &  6.84  \\
h-attn-rank \citep{Yulicheng2017}  &  20.78 &  7.38  \\
h-(gd)attn-rank \citep{Yulicheng2017}  &  21.02 &  7.51  \\
AREL \citep{wang-etal-2018-metrics} & \bf 23.02 & \bf 9.4  \\
HP \citep{wang2019}  &  21.31 &  7.44  \\
HPS \citep{wang2019}  &  21.39 &  7.75  \\
HPR \citep{wang2019}  &  21.39 &  7.61  \\
HPSR \citep{wang2019}  &  21.51 &  8.03  \\
\hline
hLSTMs (Ours) &  21.67 &  7.98  \\ 
BERT-hLSTMs (Ours) &  23.00 &  8.37  \\ 
\bottomrule
\end{tabular*}
\end{table}

\begin{table}
\caption{The comparison of the nearest similar sentences for the same sentence between the models with BERT embedding and without BERT embedding via cosine similarity.} \label{tbsentenceembed_comparison}
\begin{tabular*}{\tblwidth}{@{} LLL@{} }
\toprule
 \bf Method  & \bf label & \bf the nearest 5 points in the original space\\
\midrule
\multirow{15}{*}{hLSTMs}
 & \multirow{5}{*}{ i had a great time there. } & i had a great time on vacation last weekend. \\ 
& & i went to the beach last weekend. \\
& & it was a lot of fun. \\ 
& & i had a great time. \\ 
& & i had a great time yesterday.\\
\cline{2-3}
& \multirow{5}{*}{ we took a lot of pictures. } & we had a great time. \\ 
& & there was a lot of people there. \\ 
& & everybody was very happy. \\ 
& & everybody was having a great time. \\ 
& & the big event.  \\
\cline{2-3}
& \multirow{5}{*}{ the wedding was beautiful.} & i went to the meeting yesterday. \\
& & i went for a walk last week. \\ 
& & i went down to the beach last weekend. \\ 
& & i went to the fair last weekend. \\ 
& & the scenery was beautiful. \\
\hline
\multirow{15}{*}{BERT-hLSTMs}
 & \multirow{5}{*}{ i had a great time there. } & i had a great time. \\ 
& & i had a great time with them. \\ 
& & i had a great time on vacation. \\
& & i had a great time at the beach. \\ 
& & i had a great time at the party.  \\
\cline{2-3}
& \multirow{5}{*}{ we took a lot of pictures. } & we took lots of pictures. \\ 
& & we took a lot of pictures together.  \\ 
& & we took a lot of pictures there . \\ 
& & we took lots of photos. \\
& & we had a lot of fun.  \\
\cline{2-3}
& \multirow{5}{*}{ the wedding was beautiful. } & the ceremony was beautiful. \\ 
& & it was a lovely ceremony. \\ 
& & the concert was amazing. \\ 
& & the show was great. \\ 
& & it was a beautiful event.\\
\bottomrule
\end{tabular*}
\end{table}

\begin{table}[width=1.0\linewidth,cols=3,pos=h]
\caption{The comparison of the nearest similar words for the same word between the models with BERT embedding and without BERT embedding via cosine similarity. }\label{tbwordembed_comparison}
\begin{tabular*}{\tblwidth}{@{} LLL@{}}
\toprule
 \bf Method  & \bf label & \bf the nearest 5 points in the original space\\
\midrule
\multirow{5}{*}{hLSTMs}
 &  woman  & man, boy, girl, brother, guy \\
  &  coffee &  married, wedding, guitar, flower, camp \\
  &  enjoyed &  enjoying, enjoy, carried, liked, joined  \\
 &  were &  are, 're, 'm, be, am  \\
  &  weekend  &  week, distance, afternoon, yesterday, destination  \\
\midrule
\multirow{5}{*}{BERT-hLSTMs}
 &  woman  & person, man, lady, wife, mother \\
  &  coffee &  chocolate, cream, meat, beer, candy \\
  &  enjoyed &  enjoying, enjoy, liked, loved, worth  \\
 &  were &  are, was, is, be, been  \\
  & weekend  &  afternoon, evening, morning, summer, week  \\
\bottomrule
\end{tabular*}
\end{table}

For the text, we only chose  sentences shorter than 16 for our dataset. The <NULL> token is added to the end of the sentence when its length is shorter than 15. In addition, all punctuation is removed. Then we extract the words and sentences of the dataset to construct the word dictionary and sentence dictionary of the story descriptions, respectively. It finally yields a vocabulary of 18,000 in size for the model. Next, we embed each word and each sentence in the dictionaries with the BERT model and get the corresponding word vector and sentence vector with the dimension of 768 each. Simultaneously we construct a mask vector for each sentence description with the corresponding position of the real word in a sentence being set to be 1 while other positions with <NULL> token are 0. 

We use the story-level semantic vector as the initial LSTM input. The LSTM hidden unit has  same dimension of 768 as the story-level semantic vector. The number of time steps in the S\_LSTM layer is the number of the images in a story plus 1 while the number of time steps in the W\_LSTM layer is the size of the words in a sentence plus 1. By experimenting, We got the best result (BLEU/CIDEr score on the validation set) when we trained the model with a batch size of 16 stories, set dropout probability to 0.4 and use the Adam optimizer and set the initial learning rate to 1e-3. The system was trained on an Nvidia Titan XP GPU for 30 hours, which includes data preprocessing. 
The trained system  can produce a story for a novel sequence of images in 23 seconds.

\subsection{  Evaluation} To evaluate the performance of our proposed framework on the task of visual storytelling, we employ the automatic evaluation metrics: BLEU \citep{BLEU2002} and CIDEr \citep{CIDEr2014}. BLEU is a metric based on precision which was previously used in  machine translation to measure the similarity between the generated descriptions and ground truth. CIDEr is mainly used to evaluate the generated descriptions by conducting a Term Frequency-Inverse Document Frequency weighting for each n-gram. These measures evaluate the model by computing a score that indicates the correlation between the validation results and the ground truth descriptions. Higher score represents better performance. We run experiments by using the test set.

\subsubsection{Baselines}
\noindent {\bf Automatic Evaluation.}~~
We compare our model with five baselines in Table~\ref{tbcomparison}: 
1) A simple enc-dec model using a CNN-RNN architecture, a variant of \citep{Vinyals2014Show}, which is successful in image captioning. We extended it to capture sequences rather than still images. The encoder exploits CNN to extract the visual features of sequential images.
The sequential images are converted to CNN features as the initial hidden state for story generation. 
The decoder is a RNN architecture which decodes the hidden state of sequential images to the output story text. 
2) An enc-attn-dec model using a CNN-LSTM architecture with attention mechanism, a variant from \citep{Xu2015Show}, which is another successful framework in image captioning.  It has an attention mechanism taking into account the visual features at each time step in the decoder. Again, we extended it to encode sequential images rather than individual images.
3)~A h-(gd)attn-rank model \citep{Yulicheng2017} which uses the Gated Recurrent Unit (GRU) as both an encoder for visual feature extraction and a decoder for story generation. It also selects photos from the album, using attention, but this is something we do not tackle, so is not relevant for the comparison here.
4) The AREL model \citep{wang-etal-2018-metrics} which is an Adversarial Reward Learning framework to learn an implicit reward function and optimize policy search with the learned reward function. The AREL framework  has two main modules: a policy model with a CNN-RNN architecture and a reward model with a CNN-based architecture.
5) The HP model, and a number of variants \citep{wang2019}, which is an encoder-decoder-reconstructor architecture for the album storytelling. In this model, the encoder, decoder and reconstructor are implemented by the Gated Recurrent Unit (GRU) and a bidirectional GRU (Bi-GRU).

The main innovations of our framework are the hierarchical LSTMs which learn relationships among sentences as well as words, and in addition the use of BERT embedding (as opposed to e.g. Word2vec). 
For this reason we mostly compare with baselines which are similar but for these additions. For example \citet{Vinyals2014Show} is the basic encoder-decoder without attention, \citet{Xu2015Show} has attention, and \citet{Yulicheng2017} has a  sophisticated hierarchical structure, with separate RNNs for encoding images and for generating text. We did not compare with the works using intermediate data (see Sec.~\ref{interm}). Systems trained with reinforcement learning (see Sec.~\ref{RL}) generally perform better, and we did include one of these in our table \citep{wang-etal-2018-metrics}.

\noindent {\bf Human Evaluation.}~~
 We also conduct human evaluation to further examine the quality of the generated stories from several representative models by three annotators who are researchers in computational linguistics. The evaluation criteria contains the following aspects: relevance, coherence and expressiveness. We randomly sample 150 items from test data, each item including the same sequence of images and the corresponding stories generated by different models. Three annotators were assigned to perform a pairwise comparison and choose the better one from the two stories based on the three criteria(relevance, coherence and expressiveness). When the annotator thinks two stories for the same sequence of images are equal in quality on some criterion he can remain neutral. To make a fair comparison we randomly shufﬂe the order of the stories for each task. 
 
\begin{figure}[pos=!ht]
\centering
\includegraphics[scale=0.85]{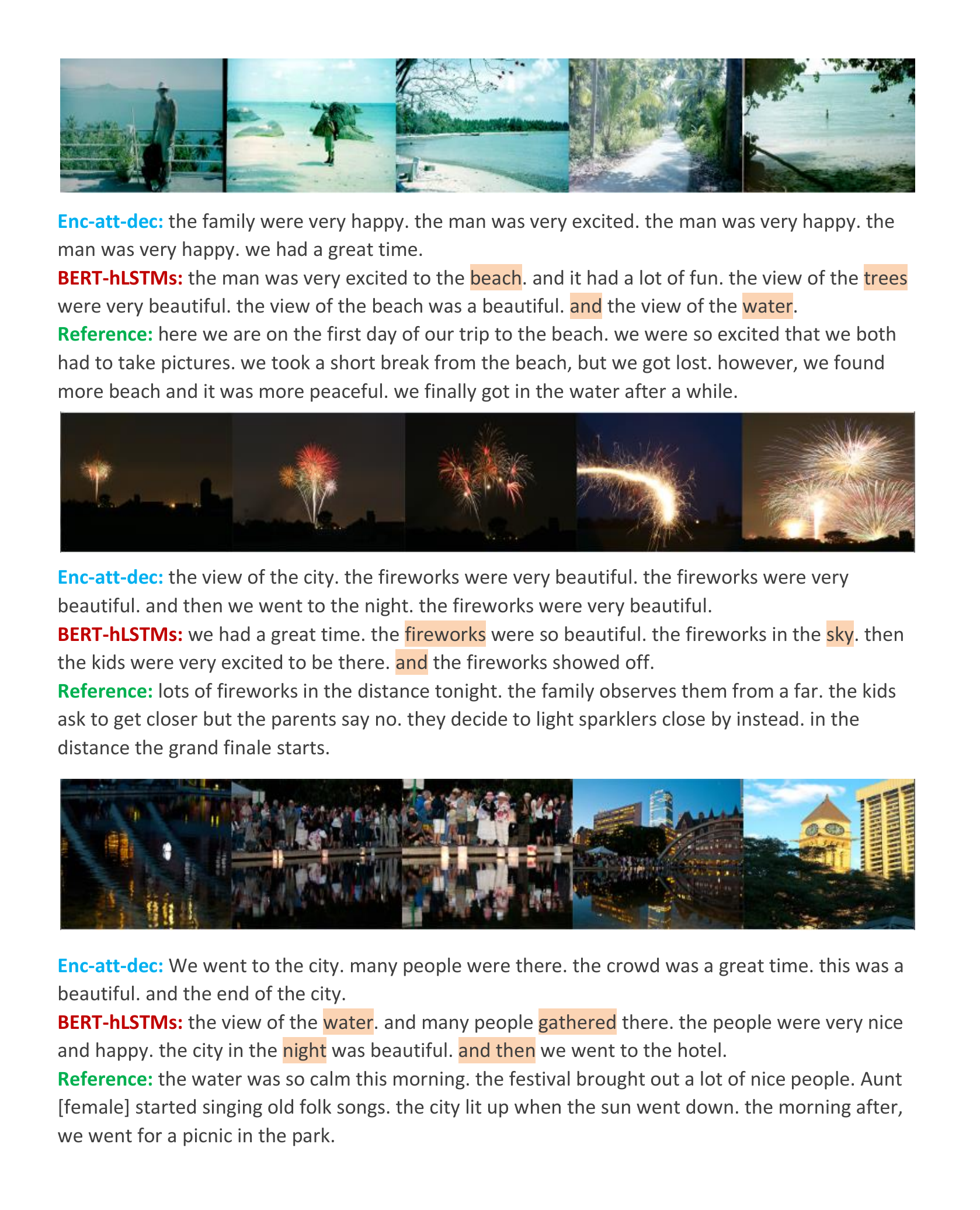}
\caption{Examples of generated stories by enc-att-dec model (variant of \citet{Xu2015Show}, extended for image sequences) in color blue and our BERT-hLSTMs model in color red, as well as the reference in color green. We highlighted certain words illustrating that our model has learnt words appropriate to construct the story.
}
\label{FIG:story example}
\end{figure}

\subsubsection{Results and Discussion}

In Table~\ref{tbcomparison}  we compared two versions of our own model, hLSTMs and BERT-hLSTMs, against other baselines, to explore the role of different components. We can see that hLSTMs without BERT embedding already achieves better performance than most of the other baselines across the metrics BLEU and CIDEr. This illustrates that hLSTMs is the most important component of our framework that contributes to its strong performance, since it enables the model to learn the dependence between sentences. In BERT-hLSTMs, the BERT component brings an additional performance improvement because it captures the more relevant and expressive semantic information between sentences during generation. Specifically, our BERT-hLSTMs model achieves  better results in the metrics with 23\% BLEU and 8.37\% CIDEr. The performance is improved by 2.49\% BLEU, 0.34\% CIDEr on the VIST dataset, compared with HPSR.

To analyse the contribution of BERT in our model, we also present some sample qualitative results from our model with and without BERT embedding, comparing the two models by looking at the similarity of their vector spaces for words and sentences.
In Table~\ref{tbsentenceembed_comparison}, we show the nearest 5 sentences of a given sentence in the original space by computing their cosine similarity. We can see that the results of the sentence embedding with BERT are obviously better than the ones without BERT. For instance, unlike the results of the second sentence without BERT, which are rarely close to the sentence  ``we took a lot of pictures'', the results with BERT are almost all similar to  ``we took a lot of pictures''. For the third sentence, although the key word  ``wedding'' was not produced by either model, our model with BERT learns the key words  ``ceremony'' ,``event ''and  ``concert'' which have a stronger correlation to  ``wedding'' than the words  ``meeting'',  ``walk'', and  ``beach'' learned by the non-BERT hLSTMs model. Similarily, from Table~\ref{tbwordembed_comparison}, we can see that the results of the words embedded with BERT are obviously better than the ones without BERT. From the nearest 5 words close to the word  ``woman'', we observe that three words  ``lady'',  ``mother'' and  ``wife'' learned by the BERT-hLSTMs model represent femininity and are very close to the word  ``woman'' while only one female word is learned by the non-BERT hLSTMs model and the other four ones are all male words like  ``man'' and ``brother''. For another word,  ``weekend'', our model can learn the nearest 5 words which are all related to time such as  ``afternoon'' and  ``summer'' while the model without BERT learns the words  ``destination'' and  ``distance'' not related to time. In conclusion, the BERT-hLSTMs model significantly outperforms the baseline non-BERT hLSTMs model.

We then compare our results with the baseline (enc-att-dec) \citep{Xu2015Show}  and ground truth on the VIST dataset. The results are listed in Figure~\ref{FIG:story example}. It can be observed that our proposed model achieves more relevant and coherent semantic information than the baseline on the VIST dataset because there are more repeated sentences in the enc-att-dec model and our model not only generates richer descriptions but also can automatically learn the conjunction  ``and'' or  ``and then''.

From the samples as shown in Figure~\ref{FIG:story example}, we see that although our BERT-hLSTMs output is not as human-like as the reference text, it exhibits a higher correlation (than baseline) with appropriate sentence-level semantics. For instance, in the first sample, our model learns the words  ``beach'',  ``trees'' and  ``water'' related to the images while the enc-att-dec model generates more repeated sentences  ``the man was very happy'' which have less relevance to the corresponding images. In the second sample, the proposed model learns to give an appropriate final  ``the fireworks showed off'' for the fifth image while not generating the same description  ``the fireworks were very beautiful'' as the previous sentence, although there are fireworks in all the images. Therefore, we conclude that the combination of BERT and hLSTMs enables the model to learn implicit sentence-level dependencies. 
We did also find some errors where the system hallucinated; for example, it sometimes produced ``beach'' and ``camera'' when these were not in the images, and failed to produce  ``meat patties'' and  ``firecrackers'' where these were visible. This tends to happen for words with low frequency in the dataset. 
Looking across more results we were able to see that the system works well on scenery images, but works poorly on  images with  many objects present, such as a market or football stadium. It is difficult for the system to know which objects might be important to the story. To get closer to human-level stories a system would need to have models of activities, such as weddings, hiking trips, eating out. The number of models would need to be large and each model would need to be very complex and elaborate, to capture many variations. This is part of the commonsense knowledge problem.
In general, despite improving on the baseline scores, our model is still far from human performance on the task of visual storytelling. The model typically captures part of the semantic information from the input image but still has room to improve.

\begin{figure}
	\centering
		\includegraphics[width=10cm]{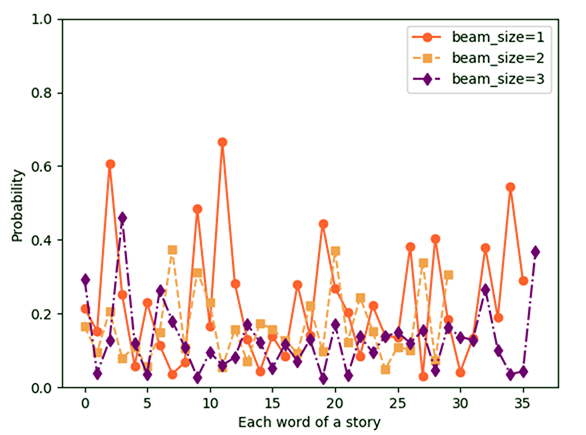}
	\caption{Comparison of the test results with different beam sizes. It shows each word probability in the generated story with beam size being 1,2 and 3.}
	\label{FIG:story}
\end{figure}

For test data, our model produces the story for sequential images by employing the beam search strategy which has had successful application in machine translation \citep{RNN2014} and image captioning previously \citep{Xu2015Show}. In the experiment, we observe that using a beam size of 1 significantly improves the story quality with a 3.95\% gain in BLEU score. Figure~\ref{FIG:story} shows the results of the generated story using the different beam sizes.

\begin{table}[width=.9\linewidth,cols=7,pos=h]
\caption{ Pairwise human comparison between BERT-hLSTMs and two methods on three aspects: Relevance, Coherence and Expressiveness. For each pairwise comparison, the first two columns show the average percentage of times that annotators preferred the model output and the third column shows the average percentage where annotators considered the two model outputs  equal in quality. Here, the Fleiss’ kappa coefficient greatly exceeds 0.5, ensuring agreement between multiple annotators. } \label{tbhuman_comparison}
\begin{tabular*}{\tblwidth}{@{} L|CCCC|CCCC@{} }
\toprule
    &  enc-att-dec  & BERT-hLSTMs  & Tie&p-value  & AREL & BERT-hLSTMs & Tie&p-value \\
\midrule
Relevance  & 27.3\% & 63.2\% & 9.5\% &.0023& 43.3\% & 46.7\% & 10.0\%&.37 \\
Coherence  & 24.7\% & 66.6\% & 8.7\% &.0002& 40.2\% & 46.1\% & 13.7\%&.12\\
Expressiveness  & 19.4\% & 72.6\% & 8.0\% &.0011& 38.9\% & 50.5\% & 10.6\%&.025\\
\bottomrule
\end{tabular*}
\end{table}

The human evaluation is shown in Table~\ref{tbhuman_comparison}, which reports the pairwise comparison between BERT-hLSTMs and the other two methods: enc-att-dec and AREL, which are the basic model from the variant of \citet{Xu2015Show} and the strongest baseline model on automatic evaluation, respectively.
A paired t-test was run on the participant percentages for preferring each model, and the two-tailed p-value is given in the table. The p-values show that the annotators preference for our approach over enc-att-dec is significant across all aspects (Relevance, Coherence and Expressiveness).
However, compared with AREL our approach was only significantly better for Expressiveness, even though AREL obtains more competitive scores on automatic metrics by using adversarial reward learning. 
We suspect that the introduction of BERT is able to enrich the semantic information of the generated story, which results in the significant improvement in the expressiveness score. 

\section{Conclusion}

In this paper, we firstly introduced a novel end-to-end BERT-hLSTMs framework, which integrates CNN, BERT and  hierarchical LSTMs to automatically generate coherent descriptions for sequential images. 
Unlike other models for visual storytelling whose decoders generate word-level descriptions, our proposed method combines sentence-level and word-level semantic information using BERT-hLSTMs. 
Our proposed method has a more straightforward network structure, has fewer training parameters and fuses the sentence-level and word-level semantic information. 
It enables our model to learn the relations between sentences (i.e. beyond the word-level) and generate more coherent descriptions.
Extensive experiments show that our BERT-hLSTMs model achieves better performance on the VIST dataset than many baselines, showing the effectiveness of using BERT embedding and the hLSTMs for sentences and words. 
To gain further performance future work could combine our hLSTMs with ideas in the best performing related works, for example, to augment with intermediate data, e.g., a scene graph \citep{wangaaai2020}, and train with reinforcement learning \citep{Huang2019}.

\bibliographystyle{cas-model2-names}
\bibliography{cas-refs}

\begin{thebibliography}{52}
\expandafter\ifx\csname natexlab\endcsname\relax\def\natexlab#1{#1}\fi
\providecommand{\url}[1]{\texttt{#1}}
\providecommand{\href}[2]{#2}
\providecommand{\path}[1]{#1}
\providecommand{\DOIprefix}{doi:}
\providecommand{\ArXivprefix}{arXiv:}
\providecommand{\URLprefix}{URL: }
\providecommand{\Pubmedprefix}{pmid:}
\providecommand{\doi}[1]{\href{http://dx.doi.org/#1}{\path{#1}}}
\providecommand{\Pubmed}[1]{\href{pmid:#1}{\path{#1}}}
\providecommand{\bibinfo}[2]{#2}
\ifx\xfnm\relax \def\xfnm[#1]{\unskip,\space#1}\fi
\bibitem[{Agrawal et~al.(2016)Agrawal, Chandrasekaran, Batra, Parikh and
  Bansal}]{SortStory2016}
\bibinfo{author}{Agrawal, H.}, \bibinfo{author}{Chandrasekaran, A.},
  \bibinfo{author}{Batra, D.}, \bibinfo{author}{Parikh, D.},
  \bibinfo{author}{Bansal, M.}, \bibinfo{year}{2016}.
\newblock \bibinfo{title}{Sort story: Sorting jumbled images and captions into
  stories}, in: \bibinfo{booktitle}{Proceedings of the 2016 Conference on
  Empirical Methods in Natural Language Processing},
  \bibinfo{publisher}{Association for Computational Linguistics},
  \bibinfo{address}{Austin, Texas}. pp. \bibinfo{pages}{925--931}.
\newblock \URLprefix \url{https://www.aclweb.org/anthology/D16-1091},
  \DOIprefix\doi{10.18653/v1/D16-1091}.
\bibitem[{Bahdanau et~al.(2015)Bahdanau, Cho and Bengio}]{Bahdanau2014Neural}
\bibinfo{author}{Bahdanau, D.}, \bibinfo{author}{Cho, K.},
  \bibinfo{author}{Bengio, Y.}, \bibinfo{year}{2015}.
\newblock \bibinfo{title}{Neural machine translation by jointly learning to
  align and translate}, in: \bibinfo{editor}{Bengio, Y.},
  \bibinfo{editor}{LeCun, Y.} (Eds.), \bibinfo{booktitle}{3rd International
  Conference on Learning Representations, {ICLR} 2015, San Diego, CA, USA, May
  7-9, 2015, Conference Track Proceedings}.
\bibitem[{Basha et~al.(2012)Basha, Moses and Avidan}]{Basha2014}
\bibinfo{author}{Basha, T.}, \bibinfo{author}{Moses, Y.},
  \bibinfo{author}{Avidan, S.}, \bibinfo{year}{2012}.
\newblock \bibinfo{title}{Photo sequencing}, in: \bibinfo{editor}{Fitzgibbon,
  A.}, \bibinfo{editor}{Lazebnik, S.}, \bibinfo{editor}{Perona, P.},
  \bibinfo{editor}{Sato, Y.}, \bibinfo{editor}{Schmid, C.} (Eds.),
  \bibinfo{booktitle}{Computer Vision -- ECCV 2012},
  \bibinfo{publisher}{Springer Berlin Heidelberg}, \bibinfo{address}{Berlin,
  Heidelberg}. pp. \bibinfo{pages}{654--667}.
\bibitem[{Bosselut et~al.(2016)Bosselut, Chen, Warren, Hajishirzi and
  Choi}]{Bosselut2016}
\bibinfo{author}{Bosselut, A.}, \bibinfo{author}{Chen, J.},
  \bibinfo{author}{Warren, D.}, \bibinfo{author}{Hajishirzi, H.},
  \bibinfo{author}{Choi, Y.}, \bibinfo{year}{2016}.
\newblock \bibinfo{title}{Learning prototypical event structure from photo
  albums}, in: \bibinfo{booktitle}{Proceedings of the 54th Annual Meeting of
  the Association for Computational Linguistics (Volume 1: Long Papers)},
  \bibinfo{publisher}{Association for Computational Linguistics},
  \bibinfo{address}{Berlin, Germany}. pp. \bibinfo{pages}{1769--1779}.
\newblock \URLprefix \url{https://www.aclweb.org/anthology/P16-1167},
  \DOIprefix\doi{10.18653/v1/P16-1167}.
\bibitem[{Cho et~al.(2014)Cho, {van Merrienboer}, Gulcehre, Bougares, Schwenk
  and Bengio}]{Cho2014Learning}
\bibinfo{author}{Cho, K.}, \bibinfo{author}{{van Merrienboer}, B.},
  \bibinfo{author}{Gulcehre, C.}, \bibinfo{author}{Bougares, F.},
  \bibinfo{author}{Schwenk, H.}, \bibinfo{author}{Bengio, Y.},
  \bibinfo{year}{2014}.
\newblock \bibinfo{title}{Learning phrase representations using rnn
  encoder-decoder for statistical machine translation}, in:
  \bibinfo{booktitle}{Conference on Empirical Methods in Natural Language
  Processing (EMNLP 2014)}.
\bibitem[{{Choi} et~al.(2016){Choi}, {Oh} and {Kweon}}]{Choi2016}
\bibinfo{author}{{Choi}, J.}, \bibinfo{author}{{Oh}, T.},
  \bibinfo{author}{{Kweon}, I.S.}, \bibinfo{year}{2016}.
\newblock \bibinfo{title}{Video-story composition via plot analysis}, in:
  \bibinfo{booktitle}{2016 IEEE Conference on Computer Vision and Pattern
  Recognition (CVPR)}, pp. \bibinfo{pages}{3122--3130}.
\bibitem[{Choi et~al.(2017)Choi, Oh and Kweon}]{Choi2017}
\bibinfo{author}{Choi, J.}, \bibinfo{author}{Oh, T.H.}, \bibinfo{author}{Kweon,
  I.S.}, \bibinfo{year}{2017}.
\newblock \bibinfo{title}{Textually customized video summaries}.
\newblock \bibinfo{journal}{arXiv preprint arXiv:1702.01528} .
\bibitem[{Devlin et~al.(2019)Devlin, Chang, Lee and Toutanova}]{BERT2018}
\bibinfo{author}{Devlin, J.}, \bibinfo{author}{Chang, M.W.},
  \bibinfo{author}{Lee, K.}, \bibinfo{author}{Toutanova, K.},
  \bibinfo{year}{2019}.
\newblock \bibinfo{title}{{BERT}: Pre-training of deep bidirectional
  transformers for language understanding}, in: \bibinfo{booktitle}{Proceedings
  of the 2019 Conference of the North {A}merican Chapter of the Association for
  Computational Linguistics: Human Language Technologies, Volume 1 (Long and
  Short Papers)}, \bibinfo{publisher}{Association for Computational
  Linguistics}, \bibinfo{address}{Minneapolis, Minnesota}. pp.
  \bibinfo{pages}{4171--4186}.
\bibitem[{{Donahue} et~al.(2017){Donahue}, {Hendricks}, {Rohrbach},
  {Venugopalan}, {Guadarrama}, {Saenko} and {Darrell}}]{Donahue2014}
\bibinfo{author}{{Donahue}, J.}, \bibinfo{author}{{Hendricks}, L.A.},
  \bibinfo{author}{{Rohrbach}, M.}, \bibinfo{author}{{Venugopalan}, S.},
  \bibinfo{author}{{Guadarrama}, S.}, \bibinfo{author}{{Saenko}, K.},
  \bibinfo{author}{{Darrell}, T.}, \bibinfo{year}{2017}.
\newblock \bibinfo{title}{Long-term recurrent convolutional networks for visual
  recognition and description}.
\newblock \bibinfo{journal}{IEEE Transactions on Pattern Analysis and Machine
  Intelligence} \bibinfo{volume}{39}, \bibinfo{pages}{677--691}.
\bibitem[{Gong et~al.(2014)Gong, Chao, Grauman and Sha}]{Gong2014}
\bibinfo{author}{Gong, B.}, \bibinfo{author}{Chao, W.L.},
  \bibinfo{author}{Grauman, K.}, \bibinfo{author}{Sha, F.},
  \bibinfo{year}{2014}.
\newblock \bibinfo{title}{Diverse sequential subset selection for supervised
  video summarization}, in: \bibinfo{editor}{Ghahramani, Z.},
  \bibinfo{editor}{Welling, M.}, \bibinfo{editor}{Cortes, C.},
  \bibinfo{editor}{Lawrence, N.D.}, \bibinfo{editor}{Weinberger, K.Q.} (Eds.),
  \bibinfo{booktitle}{Advances in Neural Information Processing Systems 27}.
  \bibinfo{publisher}{Curran Associates, Inc.}, pp.
  \bibinfo{pages}{2069--2077}.
\bibitem[{Huang et~al.(2019)Huang, Gan, Asli, Wu, Wang and He}]{Huang2019}
\bibinfo{author}{Huang, Q.}, \bibinfo{author}{Gan, Z.}, \bibinfo{author}{Asli,
  C.}, \bibinfo{author}{Wu, D.}, \bibinfo{author}{Wang, J.},
  \bibinfo{author}{He, X.}, \bibinfo{year}{2019}.
\newblock \bibinfo{title}{Hierarchically structured reinforcement learning for
  topically coherent visual story generation}.
\newblock \bibinfo{journal}{Proceedings of the AAAI Conference on Artificial
  Intelligence} \bibinfo{volume}{33}, \bibinfo{pages}{8465--8472}.
\newblock \DOIprefix\doi{10.1609/aaai.v33i01.33018465}.
\bibitem[{Huang et~al.(2016)Huang, Ferraro, Mostafazadeh, Misra, Agrawal,
  Devlin, Girshick, He, Kohli, Batra, Zitnick, Parikh, Vanderwende, Galley and
  Mitchell}]{Huang2016}
\bibinfo{author}{Huang, T.H.K.}, \bibinfo{author}{Ferraro, F.},
  \bibinfo{author}{Mostafazadeh, N.}, \bibinfo{author}{Misra, I.},
  \bibinfo{author}{Agrawal, A.}, \bibinfo{author}{Devlin, J.},
  \bibinfo{author}{Girshick, R.}, \bibinfo{author}{He, X.},
  \bibinfo{author}{Kohli, P.}, \bibinfo{author}{Batra, D.},
  \bibinfo{author}{Zitnick, C.L.}, \bibinfo{author}{Parikh, D.},
  \bibinfo{author}{Vanderwende, L.}, \bibinfo{author}{Galley, M.},
  \bibinfo{author}{Mitchell, M.}, \bibinfo{year}{2016}.
\newblock \bibinfo{title}{Visual storytelling}, in: \bibinfo{booktitle}{the
  Conference of the North American Chapter of the Association for Computational
  Linguistics}, pp. \bibinfo{pages}{1233--1239}.
\bibitem[{Karpathy and Li(2015)}]{Karpathy2015Deep}
\bibinfo{author}{Karpathy, A.}, \bibinfo{author}{Li, F.F.},
  \bibinfo{year}{2015}.
\newblock \bibinfo{title}{Deep visual-semantic alignments for generating image
  descriptions}, in: \bibinfo{booktitle}{Computer Vision and Pattern
  Recognition}, pp. \bibinfo{pages}{3128--3137}.
\bibitem[{Khosla et~al.(2013)Khosla, Hamid, Lin and Sundaresan}]{khosla2013}
\bibinfo{author}{Khosla, A.}, \bibinfo{author}{Hamid, R.},
  \bibinfo{author}{Lin, C.J.}, \bibinfo{author}{Sundaresan, N.},
  \bibinfo{year}{2013}.
\newblock \bibinfo{title}{Large-scale video summarization using web-image
  priors}, in: \bibinfo{booktitle}{The IEEE Conference on Computer Vision and
  Pattern Recognition (CVPR)}.
\bibitem[{{Kim} et~al.(2015){Kim}, {Seungwhan Moon} and {Sigal}}]{KimSigal2015}
\bibinfo{author}{{Kim}, G.}, \bibinfo{author}{{Seungwhan Moon}},
  \bibinfo{author}{{Sigal}, L.}, \bibinfo{year}{2015}.
\newblock \bibinfo{title}{Joint photo stream and blog post summarization and
  exploration}, in: \bibinfo{booktitle}{2015 IEEE Conference on Computer Vision
  and Pattern Recognition (CVPR)}, pp. \bibinfo{pages}{3081--3089}.
\bibitem[{Kim and Xing(2014)}]{Kim2014}
\bibinfo{author}{Kim, G.}, \bibinfo{author}{Xing, E.P.}, \bibinfo{year}{2014}.
\newblock \bibinfo{title}{Reconstructing storyline graphs for image
  recommendation from web community photos}, in: \bibinfo{booktitle}{IEEE
  Conference on Computer Vision and Pattern Recognition}, pp.
  \bibinfo{pages}{3882--3889}.
\bibitem[{Lavie and Agarwal(2007)}]{meteor}
\bibinfo{author}{Lavie, A.}, \bibinfo{author}{Agarwal, A.},
  \bibinfo{year}{2007}.
\newblock \bibinfo{title}{Meteor: An automatic metric for mt evaluation with
  high levels of correlation with human judgments}, in:
  \bibinfo{booktitle}{Proceedings of the Second Workshop on Statistical Machine
  Translation}, \bibinfo{publisher}{Association for Computational Linguistics},
  \bibinfo{address}{USA}. p. \bibinfo{pages}{228–231}.
\bibitem[{Li et~al.(2015)Li, Luong and Jurafsky}]{li-etal-2015-hierarchical}
\bibinfo{author}{Li, J.}, \bibinfo{author}{Luong, T.},
  \bibinfo{author}{Jurafsky, D.}, \bibinfo{year}{2015}.
\newblock \bibinfo{title}{A hierarchical neural autoencoder for paragraphs and
  documents}, in: \bibinfo{booktitle}{Proceedings of the 53rd Annual Meeting of
  the Association for Computational Linguistics and the 7th International Joint
  Conference on Natural Language Processing (Volume 1: Long Papers)},
  \bibinfo{publisher}{Association for Computational Linguistics},
  \bibinfo{address}{Beijing, China}. pp. \bibinfo{pages}{1106--1115}.
\bibitem[{Li et~al.(2019a)Li, Shi, Tang, Wu and Zhuang}]{li2019}
\bibinfo{author}{Li, J.}, \bibinfo{author}{Shi, H.}, \bibinfo{author}{Tang,
  S.}, \bibinfo{author}{Wu, F.}, \bibinfo{author}{Zhuang, Y.},
  \bibinfo{year}{2019}a.
\newblock \bibinfo{title}{Informative visual storytelling with cross-modal
  rules}, in: \bibinfo{booktitle}{Proceedings of the 27th ACM International
  Conference on Multimedia}, \bibinfo{publisher}{Association for Computing
  Machinery}, \bibinfo{address}{New York, NY, USA}. p.
  \bibinfo{pages}{2314–2322}.
\bibitem[{Li et~al.(2019b)Li, Li, Lin, Collinson and Mao}]{li2019stable}
\bibinfo{author}{Li, R.}, \bibinfo{author}{Li, X.}, \bibinfo{author}{Lin, C.},
  \bibinfo{author}{Collinson, M.}, \bibinfo{author}{Mao, R.},
  \bibinfo{year}{2019}b.
\newblock \bibinfo{title}{A stable variational autoencoder for text modelling},
  in: \bibinfo{booktitle}{Proceedings of the 12th International Conference on
  Natural Language Generation}, pp. \bibinfo{pages}{594--599}.
\bibitem[{Li et~al.(2019c)Li, Lin, Collinson, Li and Chen}]{li2019dual}
\bibinfo{author}{Li, R.}, \bibinfo{author}{Lin, C.},
  \bibinfo{author}{Collinson, M.}, \bibinfo{author}{Li, X.},
  \bibinfo{author}{Chen, G.}, \bibinfo{year}{2019}c.
\newblock \bibinfo{title}{A dual-attention hierarchical recurrent neural
  network for dialogue act classification}, in: \bibinfo{booktitle}{Proceedings
  of the 23rd Conference on Computational Natural Language Learning (CoNLL)},
  pp. \bibinfo{pages}{383--392}.
\bibitem[{Li and Li(2019)}]{li2019t}
\bibinfo{author}{Li, T.}, \bibinfo{author}{Li, S.}, \bibinfo{year}{2019}.
\newblock \bibinfo{title}{Incorporating textual evidence in visual
  storytelling}, in: \bibinfo{booktitle}{Proceedings of the 1st Workshop on
  Discourse Structure in Neural NLG, November 1},
  \bibinfo{publisher}{Association for Computational Linguistics}. p.
  \bibinfo{pages}{13–17}.
\bibitem[{Lu et~al.(2017)Lu, Xiong, Parikh and Socher}]{Lu16}
\bibinfo{author}{Lu, J.}, \bibinfo{author}{Xiong, C.}, \bibinfo{author}{Parikh,
  D.}, \bibinfo{author}{Socher, R.}, \bibinfo{year}{2017}.
\newblock \bibinfo{title}{Knowing when to look: Adaptive attention via a visual
  sentinel for image captioning}, in: \bibinfo{booktitle}{The IEEE Conference
  on Computer Vision and Pattern Recognition (CVPR)}.
\bibitem[{Mao et~al.(2014)Mao, Xu, Yang, Wang, Huang and Yuille}]{Mao2014a}
\bibinfo{author}{Mao, J.}, \bibinfo{author}{Xu, W.}, \bibinfo{author}{Yang,
  Y.}, \bibinfo{author}{Wang, J.}, \bibinfo{author}{Huang, Z.},
  \bibinfo{author}{Yuille, A.}, \bibinfo{year}{2014}.
\newblock \bibinfo{title}{Deep captioning with multimodal recurrent neural
  networks (m-rnn)}.
\newblock \bibinfo{journal}{arXiv preprint arXiv:1412.6632} .
\bibitem[{Miller(2005)}]{Miller2005}
\bibinfo{author}{Miller, J.}, \bibinfo{year}{2005}.
\newblock \bibinfo{title}{Storytelling evolves on the web: Case study: Exocog
  and the future of storytelling}.
\newblock \bibinfo{journal}{Interactions} \bibinfo{volume}{12},
  \bibinfo{pages}{30–47}.
\bibitem[{Nahian et~al.(2019)Nahian, Tasrin, Gandhi, Gaines and
  Harrison}]{Nahian2019}
\bibinfo{author}{Nahian, M.}, \bibinfo{author}{Tasrin, T.},
  \bibinfo{author}{Gandhi, S.}, \bibinfo{author}{Gaines, R.},
  \bibinfo{author}{Harrison, B.}, \bibinfo{year}{2019}.
\newblock \bibinfo{title}{A Hierarchical Approach for Visual Storytelling Using
  Image Description}.
\newblock pp. \bibinfo{pages}{304--317}.
\newblock \DOIprefix\doi{10.1007/978-3-030-33894-7_30}.
\bibitem[{Papineni et~al.(2002)Papineni, Roukos, Ward and Zhu}]{BLEU2002}
\bibinfo{author}{Papineni, K.}, \bibinfo{author}{Roukos, S.},
  \bibinfo{author}{Ward, T.}, \bibinfo{author}{Zhu, W.J.},
  \bibinfo{year}{2002}.
\newblock \bibinfo{title}{{B}leu: a method for automatic evaluation of machine
  translation}, in: \bibinfo{booktitle}{Proceedings of the 40th Annual Meeting
  of the Association for Computational Linguistics},
  \bibinfo{publisher}{Association for Computational Linguistics},
  \bibinfo{address}{Philadelphia, Pennsylvania, USA}. pp.
  \bibinfo{pages}{311--318}.
\newblock \URLprefix \url{https://www.aclweb.org/anthology/P02-1040},
  \DOIprefix\doi{10.3115/1073083.1073135}.
\bibitem[{Park and Kim(2015)}]{ImageStream2015}
\bibinfo{author}{Park, C.C.}, \bibinfo{author}{Kim, G.}, \bibinfo{year}{2015}.
\newblock \bibinfo{title}{Expressing an image stream with a sequence of natural
  sentences}, in: \bibinfo{booktitle}{Advances in Neural Information Processing
  Systems 28}, pp. \bibinfo{pages}{73--81}.
\bibitem[{{Pickup} et~al.(2014){Pickup}, {Pan}, {Wei}, {Shih}, {Zhang},
  {Zisserman}, {Scholkopf} and {Freeman}}]{Pickup2014}
\bibinfo{author}{{Pickup}, L.C.}, \bibinfo{author}{{Pan}, Z.},
  \bibinfo{author}{{Wei}, D.}, \bibinfo{author}{{Shih}, Y.},
  \bibinfo{author}{{Zhang}, C.}, \bibinfo{author}{{Zisserman}, A.},
  \bibinfo{author}{{Scholkopf}, B.}, \bibinfo{author}{{Freeman}, W.T.},
  \bibinfo{year}{2014}.
\newblock \bibinfo{title}{Seeing the arrow of time}, in:
  \bibinfo{booktitle}{2014 IEEE Conference on Computer Vision and Pattern
  Recognition}, pp. \bibinfo{pages}{2043--2050}.
\bibitem[{Ramanathan et~al.(2015)Ramanathan, Tang, Mori and
  Fei-Fei}]{Ramanathan2015}
\bibinfo{author}{Ramanathan, V.}, \bibinfo{author}{Tang, K.},
  \bibinfo{author}{Mori, G.}, \bibinfo{author}{Fei-Fei, L.},
  \bibinfo{year}{2015}.
\newblock \bibinfo{title}{Learning temporal embeddings for complex video
  analysis}, in: \bibinfo{booktitle}{Proceedings of the 2015 IEEE International
  Conference on Computer Vision (ICCV)}, \bibinfo{publisher}{IEEE Computer
  Society}, \bibinfo{address}{USA}. p. \bibinfo{pages}{4471–4479}.
\bibitem[{Serban et~al.(2016)Serban, Sordoni, Bengio, Courville and
  Pineau}]{Serban}
\bibinfo{author}{Serban, I.V.}, \bibinfo{author}{Sordoni, A.},
  \bibinfo{author}{Bengio, Y.}, \bibinfo{author}{Courville, A.},
  \bibinfo{author}{Pineau, J.}, \bibinfo{year}{2016}.
\newblock \bibinfo{title}{Building end-to-end dialogue systems using generative
  hierarchical neural network models}, in: \bibinfo{booktitle}{Proceedings of
  the Thirtieth AAAI Conference on Artificial Intelligence},
  \bibinfo{publisher}{AAAI Press}. p. \bibinfo{pages}{3776–3783}.
\bibitem[{Sigurdsson et~al.(2016)Sigurdsson, Chen and Gupta}]{Sigurdsson2016}
\bibinfo{author}{Sigurdsson, G.A.}, \bibinfo{author}{Chen, X.},
  \bibinfo{author}{Gupta, A.}, \bibinfo{year}{2016}.
\newblock \bibinfo{title}{Learning visual storylines with skipping recurrent
  neural networks}, in: \bibinfo{booktitle}{European Conference on Computer
  Vision}, \bibinfo{organization}{Springer}. pp. \bibinfo{pages}{71--88}.
\bibitem[{Simonyan and Zisserman(2015)}]{Simonyan2014Very}
\bibinfo{author}{Simonyan, K.}, \bibinfo{author}{Zisserman, A.},
  \bibinfo{year}{2015}.
\newblock \bibinfo{title}{Very deep convolutional networks for large-scale
  image recognition}, in: \bibinfo{booktitle}{International Conference on
  Learning Representations}.
\bibitem[{Su et~al.(2018)Su, Lin, Zhou, Dai and Lv}]{su2018generating}
\bibinfo{author}{Su, J.}, \bibinfo{author}{Lin, C.}, \bibinfo{author}{Zhou,
  M.}, \bibinfo{author}{Dai, Q.}, \bibinfo{author}{Lv, H.},
  \bibinfo{year}{2018}.
\newblock \bibinfo{title}{Generating description for sequential images with
  local-object attention conditioned on global semantic context}, in:
  \bibinfo{booktitle}{Proceedings of the Workshop on Intelligent Interactive
  Systems and Language Generation (2IS\&NLG)}, pp. \bibinfo{pages}{3--8}.
\bibitem[{Sutskever et~al.(2014)Sutskever, Vinyals and Le}]{RNN2014}
\bibinfo{author}{Sutskever, I.}, \bibinfo{author}{Vinyals, O.},
  \bibinfo{author}{Le, Q.V.}, \bibinfo{year}{2014}.
\newblock \bibinfo{title}{Sequence to sequence learning with neural networks},
  in: \bibinfo{editor}{Ghahramani, Z.}, \bibinfo{editor}{Welling, M.},
  \bibinfo{editor}{Cortes, C.}, \bibinfo{editor}{Lawrence, N.},
  \bibinfo{editor}{Weinberger, K.Q.} (Eds.), \bibinfo{booktitle}{Advances in
  Neural Information Processing Systems}, \bibinfo{publisher}{Curran
  Associates, Inc.}. pp. \bibinfo{pages}{3104--3112}.
\bibitem[{Tali et~al.(2014)Tali, Yael and Shai}]{PhotoSequencing2014}
\bibinfo{author}{Tali, D.}, \bibinfo{author}{Yael, M.}, \bibinfo{author}{Shai,
  A.}, \bibinfo{year}{2014}.
\newblock \bibinfo{title}{Photo sequencing}.
\newblock \bibinfo{journal}{International Journal of Computer Vision}
  \bibinfo{volume}{110}, \bibinfo{pages}{275--289}.
\bibitem[{Vaswani et~al.(2017)Vaswani, Shazeer, Parmar, Uszkoreit, Jones,
  Gomez, Kaiser and Polosukhin}]{transformer2017}
\bibinfo{author}{Vaswani, A.}, \bibinfo{author}{Shazeer, N.},
  \bibinfo{author}{Parmar, N.}, \bibinfo{author}{Uszkoreit, J.},
  \bibinfo{author}{Jones, L.}, \bibinfo{author}{Gomez, A.N.},
  \bibinfo{author}{Kaiser, {\L}.}, \bibinfo{author}{Polosukhin, I.},
  \bibinfo{year}{2017}.
\newblock \bibinfo{title}{Attention is all you need}, in:
  \bibinfo{booktitle}{Advances in neural information processing systems}, pp.
  \bibinfo{pages}{5998--6008}.
\bibitem[{Vedantam et~al.(2015)Vedantam, Lawrence~Zitnick and
  Parikh}]{CIDEr2014}
\bibinfo{author}{Vedantam, R.}, \bibinfo{author}{Lawrence~Zitnick, C.},
  \bibinfo{author}{Parikh, D.}, \bibinfo{year}{2015}.
\newblock \bibinfo{title}{{CIDEr}: Consensus-based image description
  evaluation}, in: \bibinfo{booktitle}{Proceedings of the IEEE conference on
  Computer Vision and Pattern Recognition}, pp. \bibinfo{pages}{4566--4575}.
\bibitem[{Venugopalan et~al.(2015a)Venugopalan, Rohrbach, Donahue, Mooney,
  Darrell and Saenko}]{Venugopalan2015Sequence}
\bibinfo{author}{Venugopalan, S.}, \bibinfo{author}{Rohrbach, M.},
  \bibinfo{author}{Donahue, J.}, \bibinfo{author}{Mooney, R.},
  \bibinfo{author}{Darrell, T.}, \bibinfo{author}{Saenko, K.},
  \bibinfo{year}{2015}a.
\newblock \bibinfo{title}{Sequence to sequence -- video to text}, in:
  \bibinfo{booktitle}{IEEE International Conference on Computer Vision}, pp.
  \bibinfo{pages}{4534--4542}.
\bibitem[{Venugopalan et~al.(2015b)Venugopalan, Xu, Donahue, Rohrbach, Mooney
  and Saenko}]{Venugopalan2014Translating}
\bibinfo{author}{Venugopalan, S.}, \bibinfo{author}{Xu, H.},
  \bibinfo{author}{Donahue, J.}, \bibinfo{author}{Rohrbach, M.},
  \bibinfo{author}{Mooney, R.}, \bibinfo{author}{Saenko, K.},
  \bibinfo{year}{2015}b.
\newblock \bibinfo{title}{Translating videos to natural language using deep
  recurrent neural networks}, in: \bibinfo{booktitle}{Proceedings of the 2015
  Conference of the North {A}merican Chapter of the Association for
  Computational Linguistics: Human Language Technologies},
  \bibinfo{publisher}{Association for Computational Linguistics},
  \bibinfo{address}{Denver, Colorado}. pp. \bibinfo{pages}{1494--1504}.
\bibitem[{Vinyals et~al.(2015)Vinyals, Toshev, Bengio and
  Erhan}]{Vinyals2014Show}
\bibinfo{author}{Vinyals, O.}, \bibinfo{author}{Toshev, A.},
  \bibinfo{author}{Bengio, S.}, \bibinfo{author}{Erhan, D.},
  \bibinfo{year}{2015}.
\newblock \bibinfo{title}{Show and tell: A neural image caption generator}, in:
  \bibinfo{booktitle}{The IEEE Conference on Computer Vision and Pattern
  Recognition (CVPR)}, pp. \bibinfo{pages}{3156--3164}.
\bibitem[{Wang et~al.(2019)Wang, Zhang, Jiang and Zhang}]{wang2019}
\bibinfo{author}{Wang, B.}, \bibinfo{author}{Zhang, W.},
  \bibinfo{author}{Jiang, W.}, \bibinfo{author}{Zhang, F.},
  \bibinfo{year}{2019}.
\newblock \bibinfo{title}{Hierarchical photo-scene encoder for album
  storytelling}.
\newblock \bibinfo{journal}{Proceedings of the AAAI Conference on Artificial
  Intelligence} \bibinfo{volume}{33}, \bibinfo{pages}{8909--8916}.
\newblock \DOIprefix\doi{10.1609/aaai.v33i01.33018909}.
\bibitem[{Wang et~al.(2020)Wang, Wei, Li, Zhang and Huang}]{wangaaai2020}
\bibinfo{author}{Wang, R.}, \bibinfo{author}{Wei, Z.}, \bibinfo{author}{Li,
  P.}, \bibinfo{author}{Zhang, Q.}, \bibinfo{author}{Huang, X.},
  \bibinfo{year}{2020}.
\newblock \bibinfo{title}{A hierarchical approach for visual storytelling from
  an image stream using scene graphs}, in: \bibinfo{booktitle}{The
  Thirty-Fourth AAAI Conference on Artificial Intelligence}.
\bibitem[{Wang et~al.(2018)Wang, Chen, Wang and Wang}]{wang-etal-2018-metrics}
\bibinfo{author}{Wang, X.}, \bibinfo{author}{Chen, W.}, \bibinfo{author}{Wang,
  Y.F.}, \bibinfo{author}{Wang, W.Y.}, \bibinfo{year}{2018}.
\newblock \bibinfo{title}{No metrics are perfect: Adversarial reward learning
  for visual storytelling}, in: \bibinfo{booktitle}{Proceedings of the 56th
  Annual Meeting of the Association for Computational Linguistics (Volume 1:
  Long Papers)}, \bibinfo{publisher}{Association for Computational
  Linguistics}, \bibinfo{address}{Melbourne, Australia}. pp.
  \bibinfo{pages}{899--909}.
\newblock \DOIprefix\doi{10.18653/v1/P18-1083}.
\bibitem[{Xu et~al.(2015)Xu, Ba, Kiros, Cho, Courville, Salakhudinov, Zemel and
  Bengio}]{Xu2015Show}
\bibinfo{author}{Xu, K.}, \bibinfo{author}{Ba, J.}, \bibinfo{author}{Kiros,
  R.}, \bibinfo{author}{Cho, K.}, \bibinfo{author}{Courville, A.},
  \bibinfo{author}{Salakhudinov, R.}, \bibinfo{author}{Zemel, R.},
  \bibinfo{author}{Bengio, Y.}, \bibinfo{year}{2015}.
\newblock \bibinfo{title}{Show, attend and tell: Neural image caption
  generation with visual attention}, in: \bibinfo{booktitle}{International
  conference on machine learning}, pp. \bibinfo{pages}{2048--2057}.
\bibitem[{Yao et~al.(2015)Yao, Torabi, Cho, Ballas, Pal, Larochelle and
  Courville}]{Yao2015Describing}
\bibinfo{author}{Yao, L.}, \bibinfo{author}{Torabi, A.}, \bibinfo{author}{Cho,
  K.}, \bibinfo{author}{Ballas, N.}, \bibinfo{author}{Pal, C.},
  \bibinfo{author}{Larochelle, H.}, \bibinfo{author}{Courville, A.},
  \bibinfo{year}{2015}.
\newblock \bibinfo{title}{Describing videos by exploiting temporal structure},
  in: \bibinfo{booktitle}{The IEEE International Conference on Computer Vision
  (ICCV)}, pp. \bibinfo{pages}{199--211}.
\bibitem[{You et~al.(2016)You, Jin, Wang, Fang and Luo}]{You2016Image}
\bibinfo{author}{You, Q.}, \bibinfo{author}{Jin, H.}, \bibinfo{author}{Wang,
  Z.}, \bibinfo{author}{Fang, C.}, \bibinfo{author}{Luo, J.},
  \bibinfo{year}{2016}.
\newblock \bibinfo{title}{Image captioning with semantic attention}, in:
  \bibinfo{booktitle}{The IEEE Conference on Computer Vision and Pattern
  Recognition (CVPR)}, pp. \bibinfo{pages}{4651--4659}.
\bibitem[{Yu et~al.(2017)Yu, Bansal and Berg}]{Yulicheng2017}
\bibinfo{author}{Yu, L.}, \bibinfo{author}{Bansal, M.}, \bibinfo{author}{Berg,
  T.L.}, \bibinfo{year}{2017}.
\newblock \bibinfo{title}{Hierarchically-attentive {RNN} for album
  summarization and storytelling}, in: \bibinfo{booktitle}{Proceedings of the
  2017 Conference on Empirical Methods in Natural Language Processing}, pp.
  \bibinfo{pages}{966--971}.
\newblock \DOIprefix\doi{10.18653/v1/D17-1101}.
\bibitem[{Zhang et~al.(2020)Zhang, Hu and Sha}]{Zhang2020}
\bibinfo{author}{Zhang, B.}, \bibinfo{author}{Hu, H.}, \bibinfo{author}{Sha,
  F.}, \bibinfo{year}{2020}.
\newblock \bibinfo{title}{Visual storytelling via predicting anchor word
  embeddings in the stories}.
\newblock \bibinfo{journal}{CoRR} \bibinfo{volume}{abs/2001.04541}.
\newblock \href{http://arxiv.org/abs/2001.04541}{\tt arXiv:2001.04541}.
\bibitem[{Zhang et~al.(2016)Zhang, Chao, Sha and Grauman}]{Zhang2016}
\bibinfo{author}{Zhang, K.}, \bibinfo{author}{Chao, W.L.},
  \bibinfo{author}{Sha, F.}, \bibinfo{author}{Grauman, K.},
  \bibinfo{year}{2016}.
\newblock \bibinfo{title}{Summary transfer: Exemplar-based subset selection for
  video summarization}, in: \bibinfo{booktitle}{The IEEE Conference on Computer
  Vision and Pattern Recognition (CVPR)}.
\bibitem[{Zhang et~al.(2018)Zhang, Liu and Song}]{zhang2018sentencestate}
\bibinfo{author}{Zhang, Y.}, \bibinfo{author}{Liu, Q.}, \bibinfo{author}{Song,
  L.}, \bibinfo{year}{2018}.
\newblock \bibinfo{title}{Sentence-state lstm for text representation}.
\newblock \href{http://arxiv.org/abs/1805.02474}{\tt arXiv:1805.02474}.
\bibitem[{Zhu et~al.(2015)Zhu, Kiros, Zemel, Salakhutdinov, Urtasun, Torralba
  and Fidler}]{Zhu2015Aligning}
\bibinfo{author}{Zhu, Y.}, \bibinfo{author}{Kiros, R.}, \bibinfo{author}{Zemel,
  R.}, \bibinfo{author}{Salakhutdinov, R.}, \bibinfo{author}{Urtasun, R.},
  \bibinfo{author}{Torralba, A.}, \bibinfo{author}{Fidler, S.},
  \bibinfo{year}{2015}.
\newblock \bibinfo{title}{Aligning books and movies: Towards story-like visual
  explanations by watching movies and reading books}, in:
  \bibinfo{booktitle}{The IEEE International Conference on Computer Vision
  (ICCV)}, pp. \bibinfo{pages}{19--27}.

\end{thebibliography}

\end{document}